# Deep Learning from Parametrically Generated Virtual Buildings for Real-World Object Recognition

Mohammad Alawadhi[a], Wei Yan[b]

[a]*Kuwait University, Department of Architecture, Kuwait*
[b]*Texas A&M University, Department of Architecture, Texas, United States*

**Abstract**

We study the use of parametric building information modeling (BIM) to automatically generate training data for teaching artificial neural networks (ANNs) to recognize building objects in photos. Teaching artificial intelligence (AI) machines to detect building objects in photos and videos is the foundation toward achieving AI-assisted semantic 3D reconstruction of buildings, which would be useful for various applications including surveys, documentation, and performance modeling. However, there exists the challenge of acquiring enough training data for machine learning, and this data is typically manually curated and annotated by people—that is, unless a computer machine can generate high-quality data to train itself for a certain task. In that vein, we trained ANNs solely on realistic computer-generated images of 3D BIM models—which were parametrically and automatically generated using the BIMGenE program developed for this research. The ANN training result demonstrated generalizability and good semantic segmentation on a test case as well as arbitrary photos of buildings that are outside the range of the generated training data, which is significant for the future of training AI with generated data for solving real-world architectural problems.

## 1. Introduction

Interest in artificial intelligence (AI) for the fields of computational design and architecture, engineering, and construction (AEC) is not recent and began soon after the birth of AI as a field with previous work in the early 1970s [1,2]. Yet recent trends in AI research will have a significant impact on the future of AEC. The total number of AI-in-the-AEC research publications since the 1970s doubled in the latter half of the 2010s alone [3]. This explosive trend in research coincides with recent breakthroughs in deep artificial neural networks (ANNs) during the same decade, and the establishment of deep learning as a revolutionary field in computer science and beyond.

Deep learning has become the state-of-the-art form of machine learning, and deep ANNs have excelled in computer vision-based object recognition tasks. One important task that is relevant to AEC is to teach ANNs to categorize exterior building objects from real-world images (photos and videos). This would be useful for various 3D reconstruction applications when used in conjunction with UAV-assisted reality capture (i.e., 3D scanning) methods, for example, to automate scan-to-building information modeling (BIM) and to gather data for building energy modeling (BEM) for green architecture.

ANNs require training by presenting them with examples that identify a pattern within a problem. Their limitation is requiring the availability of large, high-quality, and annotated (or labeled) training datasets to be trained adequately. A suggested solution for training data acquisition is training with both BIM and computer-generated imagery (CGI)—ubiquitous concepts in AEC—to automatically synthesize training datasets for deep learning to recognize building objects in photos [4]. This can provide a new solution for creating basic 3D BIM models using multi-view photos with detected building objects [4]. However, previous work on CGI as training data come with a major caveat: Training data was acquired from manually created 3D models and limited to data that is already available. In cases where training data does not exist, there are no perceived benefits in using manually created and annotated BIM and CGI over human-annotated photos for acquiring training data, since both acquisitions are time-consuming. Where neither human-annotated data nor 3D data is available, another benefit of using a CGI or BIM platform is the ability to integrate a data synthesis pipeline with generative computation methods.

A large potential for AI is deep learning from generated 3D objects and environments. Therefore, this work presents the Building Information Model Generation Environment (BIMGenE) program that parametrically generates data to train ANNs for building object recognition. The methods and innovations of this program are described in this paper. The 3D BIM models generated from BIMGenE were automatically batch-rendered and processed as training data which was used to train ANNs that were fine-tuned on a test case, after which ANNs were tested and evaluated on arbitrary photos of buildings.

Two subtypes of ANNs were used in the deep learning experiments with the generated training data. These are convolutional neural network (CNN) and generative adversarial network (GAN). CNNs, which are inspired by the visual cortex, are the most commonly used ANN for object recognition, particularly for semantic segmentation (i.e., segmenting an input image into color-labeled objects). GANs are a newer technology based on implementing two ANNs (e.g., two CNNs) competing in a zero-sum game. GANs are used for semantic segmentation in addition to other image-based learning tasks. A GAN model was the main ANN type that was used to train with the generated parametric BIM data (conventional CNNs were also used to compare).

The semantic segmentation evaluation results of a parametric BIM-trained ANN achieved 89.64% average accuracy and 0.517 mean IoU (mIoU) on the test case, which is similar to the results of a benchmark ANN trained with manually created synthetic data [4]. Also, it achieved above 80% accuracy and above 0.5 mIoU on both hand-picked and randomly sampled sets of arbitrary photos, which are better scores than the benchmark ANN that is tested on the same arbitrary photosets.

*1.1. Literature Review*

Building facade labeling is a computer vision problem and a key component toward the 3D reconstruction of street scenes [5]. [5] conducted an early study to train a semantic segmentation CNN for building facades using a pre-existing training dataset, which is a late application of technology when compared to previous problems (e.g., indoor scene understanding). ANNs for pixel-wise semantic segmentation are technologies that are useful for 3D reconstruction that witnessed rapid growth—which can be used to create semantically labeled urban models by projecting ANN predictions from photos onto a 3D-scanned urban model [6].

A limitation is that most training datasets for semantic segmentation are human-annotated, which is labor-intensive and time-consuming. Using synthetic data (if available) is a suggested alternative to prepare annotated training datasets. BIM data (in contrast to 3D CAD data) can provide intelligent building objects to automate the annotation process. For example, [7] applied labeling through semantic segmentation as a key step toward achieving automatic as-built BIM model creation and trained an ANN using a combination of synthetic point clouds generated using BIM models, and real point clouds of scanned rooms, but found that synthetic point clouds were not sufficient to replace real point clouds. Further, most other attempts in testing BIM-trained ANNs were limited to the digital realm. More research was needed to study the feasibility of training with only synthetic BIM-based data for real-world scenarios, and to experiment with emerging ANN technologies (e.g., GANs).

Also, most of the work applying ANNs to the built environment was concerned with indoor scene understanding and reconstruction [7–12] as opposed to building exteriors. This is due to that previous applications of building exterior reconstruction were thought of for computer games and entertainment [5,6,13,14]. There is a lack of research on using deep learning from BIM to characterize building exteriors for AEC applications. However, with the growth of technology, 3D reconstruction of buildings has a wide range of applications [6], including environmental simulations [15]. [16] conducted a review on machine learning applications on the building life cycle and determined the importance of semantic segmentation of building facades to extract window-to-wall ratio, material type, etc. for building energy modeling (BEM). Therefore, there is a need to research approaches for training ANNs with BIM to characterize building exteriors—as a step toward automating 3D reconstruction for AEC.

Past work indicated that building object recognition can help automate BIM and BEM from existing buildings [4,7,17–21]. This problem lies within the wider challenge of extracting useful building semantics from multi-view photos and 3D-scanned models acquired from site surveys which lack useful information beyond geometry and color data. Extracting this information—e.g., wall, window, and roof objects within the geometry—is needed to conduct building performance simulations for green architecture and other virtual modeling tasks that require building semantics [21]. Until the early 2010s, methods relied on traditional machine learning algorithms [22], which were based on programming step-by-step instructions—a complex solution for a complex problem. With recent state-of-the-art ANNs, there remained the question if AI trained with realistically rendered BIM data (which inherently provide building object semantics) can help bridge that challenge.

Synthetic data in the form of photorealistic CGI can be a viable replacement for real photos when training ANNs to recognize objects especially if the CGI data is simulated to be very similar to photographic data [4,11,23–25]. [4] presented methods to use photorealistic CGI and BIM-based color-ID renderings for AI training. However, there is a limitation in using manually created synthetic data. Often, BIM and CGI require specialists in 3D, while human annotation can be crowd-sourced rapidly using data labeling services [26]. In general, both types of data acquisitions (manually created synthetic data and human-annotated data) are laborious and time-consuming. Alternatively, an advantage of using CGI or BIM is the ability to apply generative computation methods (i.e., training with generated data). There is a lack of previous work in training ANNs with automatically generated virtual buildings for recognizing real-world building objects.

*1.2. Parametrically Generated Buildings for Teaching AI*

An unexplored potential for AI in AEC is in deep learning from automatically generated data. 3D objects can be generated through parametric or procedural methods [27,28]. As evident in many extant virtual environments created for computer games, procedurally generated content can have photorealistic qualities—hyperreal, even. A parametric BIM solution for generating BIM models would require predefining a set of rules. To create training data synthesis these rules must produce a large enough solution space that approximates the description of many real buildings, so that a trained ANN can correctly apply the knowledge learned from synthetic data to photographed buildings. Fortunately, most architectural designs follow certain cognitive logic or rules. Human knowledge of what a typical building should look like can be applied as rules in a computer program. These rules can be parameters for intelligent BIM elements, such as: floor outlines; number of floors; dimensional constraints; object materials; window types; architectural details; etc.

The automated generation of coherent buildings using a finite set of parametric values is an idea that harkens back to *A Pattern Language* by [29]. BIMGenE was developed following a similar idea; however, the concept of patterns is replaced with parameters. A finite list of primary parameters was established as the fundamental building-blocks to describe a building. These were understood as the priori of basic architectural elements that when in concert form a coherent building. Furthermore, varying the parameters' values would produce different typologies in design.

## 2. Methods

This section describes the process to parametrically generate 3D BIM models in BIMGenE (Section 2.1) and the CGI data synthesis framework (beginning from Section 2.2) to generate the ANN training datasets. The data synthesis framework is based on previous work that developed a methodology for training data synthesis using BIM and photorealistic rendering [4].

*2.1. 3D BIM Model Generation*

The BIMGenE program was developed to provide a generative 3D modeling solution that applies BIM with parametric techniques. BIMGenE automatically generates 3D BIM models using a number of parametric controls. The BIM data of intelligent objects (wall, window, etc.) contained in the generated 3D models were used to generate color-coded ground truths for deep learning. The data synthesis framework (Section 2.2) used different views of the 3D BIM models and rendered them automatically, then the images were processed as training data to teach ANNs. Initial experiments fine-tuned the ANNs on a test case, then ANNs were tested on arbitrary photos of buildings outside the range of both the test case and the training data (Section 3).

Each 3D BIM model is automatically generated in BIMGenE based on preset input parameters where changing these parameters would generate a new model within seconds. The first process to generate a model uses standard parameters to automatically place fenestrations (windows and doors). The second uses customizable parameters to manually place fenestrations. The two methods offer flexibility to choose between a rule-based or customized approach to generate openings based on the desired characteristics of the end result training data.

Table 1 shows three examples of selected parameter values and each corresponding generated 3D BIM model. The curves for the building footprint parameter were pre-drawn as various inputs.

*2.1.1. Generating Walls, Floors, and Roof*

Walls are created following a wall reference line that is defined by a building footprint shape parameter—which can be represented by an indefinite number of polyline morphologies with straight edges, then wall polylines and edges are calculated for each floor, which are then used as inputs to generate walls, slabs, and roofs. A linear array function is applied to the building footprint shape parameter, where the array direction is a unit vector in the Z axis direction multiplied by a wall height parameter, and the array count is a number of floors parameter. BIM wall settings are specified with material, height, and thickness parameters. The material parameter defines an object from a BIM material library. BIM walls are created with the specified settings using wall polylines as the curve inputs for the reference line.

Data values and identifiers for BIM objects are stored within the program inside nested lists. All building objects are stored in a list containing sub-lists, where each sub-list stores a discrete building object. Subsequently, each of these sub-lists also store lists representing floor objects. The same is true for these floor lists, where each stores wall lists. Figure 1 shows a simple data structure example representing six discrete building objects, with one building (list 2) containing two floor lists (2:0 and 2:1) with four walls each. Under ground floor list 2:0, there are four lists where each stores a discrete wall object, and in Figure 1 there are four walls as shown with indices from 0 to 3. A similar data structure to the example (in Figure 1) is automatically formed for each generation.

**Table 1.** Examples of generated BIM models from varying BIMGenE parameters.

|  | Generation #1 | Generation #2 | Generation #3 |
|---|---|---|---|
| Building Footprint Shape Parameter | 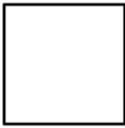 | 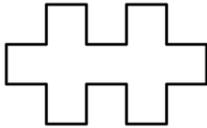 | 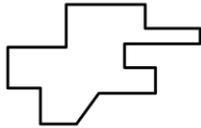 |
| Rendered BIMGenE Result | 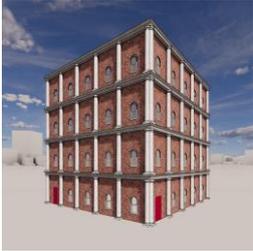 | 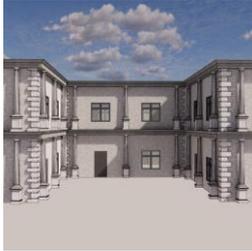 | 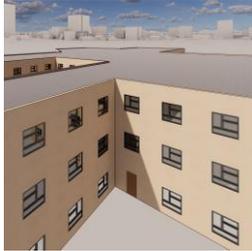 |

To generate the doors at a later step, the ground floor walls are separated from the wall data with a condition for nested list indices where the first index could be any integer; however, the second index must equal zero—because this is the index value where the ground floor walls reside in the nested list. Applying the condition of selecting all *n*:0 lists enable selecting only the ground floor walls. The conditional expression means "from all *n* indices of building lists select only floor lists which have an index of 0". Figure 2 shows the ground wall selections color-coded in the user interface (UI) after applying the condition.

A slab and a roof are generated for each curve in the wall polylines by tracing the shape of the floor plan using parameters for roof material and slab height. These curves are moved by the slab height parameter, then are offset by an arbitrary distance to avoid geometry intersections with the outer surface of the walls.

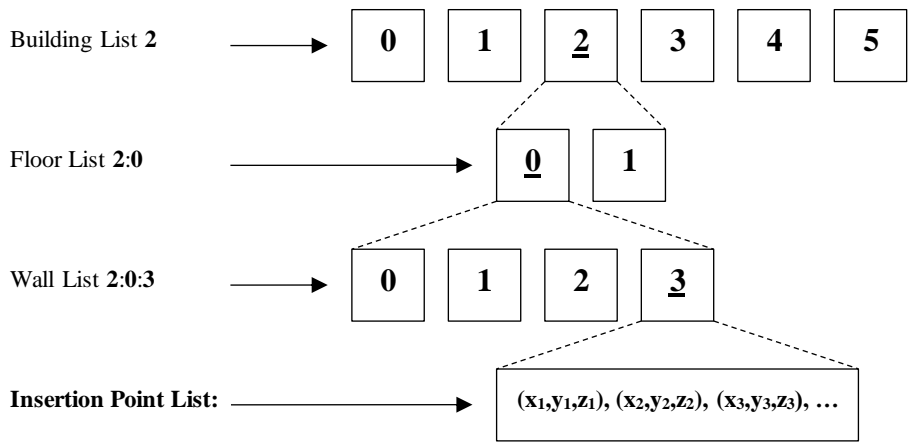

**Figure 1.** Nested list data structure where list 2:0:3—the fourth wall of the ground floor of the third building—contains a list of point values which are insertion coordinates for fenestration objects.

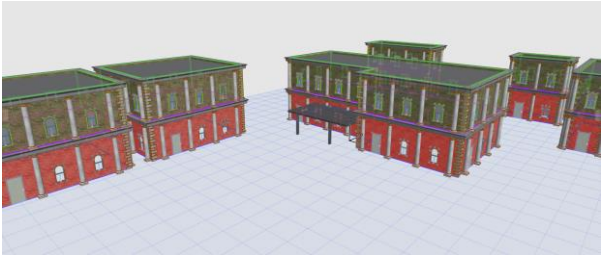

**Figure 2.** Ground walls selected in the UI (highlighted in red).

*2.1.2. Generating Fenestrations*

A BIM fenestration object is placed within a wall by specifying an insertion point on the wall reference line, the outside direction of the object, and the opening direction. An orientation point is specified relative to the insertion point and the wall reference line, where this point specifies both the outside and the opening directions of the placed fenestration. A distance between openings parameter is used where given a specified spacing length in meters, a fenestration is inserted. The modulus is calculated between the curve lengths of each wall edge and the distance between openings parameter. The remainder is used to trim the wall edges to the length of the specified spacing. This is done by extending each end of the curves to the negative direction where each extension is half the remainder. Then, the resultant curves are divided into segments and points by the specified distance in meters. The midpoints of the divided curve segments are used to specify the fenestration insertion points on the walls.

Figure 3 shows the effect of changing the distance between openings parameter which is reflected in real-time on the appearance of the generated 3D BIM model. Changing this parameter affects other elements such as the pilaster objects which are optionally placed with a Boolean using the divided wall segment points—which are the points between the insertion points.

To separate opening insertion points into window and door insertion points, it is assumed that the first point in each distinct wall element on the ground floor is for a door, and the rest of the points are for windows. Figure 1 shows how an insertion point list is stored in wall list 2:0:3. For the nested list data structure $n_1:n_2:n_3$ for the opening insertion points, applying a splitting condition and then selecting the first point in each list will separate the first insertion point of each wall on the ground floor. The conditional expression means "from all $n_1$ indices of building lists and $n_2$ indices of floor lists select only the first wall lists which have an index of 0". This will generate the door insertion points separately as highlighted in red in Figure 4. To get the window insertion points, a set difference is calculated between the opening insertion points and the door insertion points.

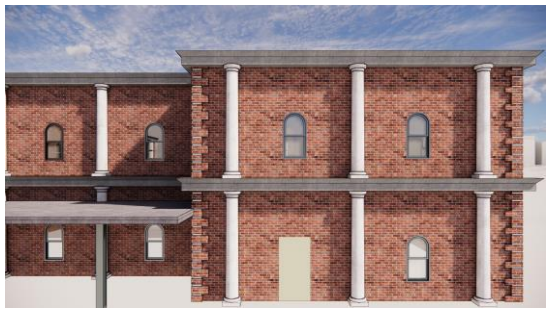

distance between openings = 4 m

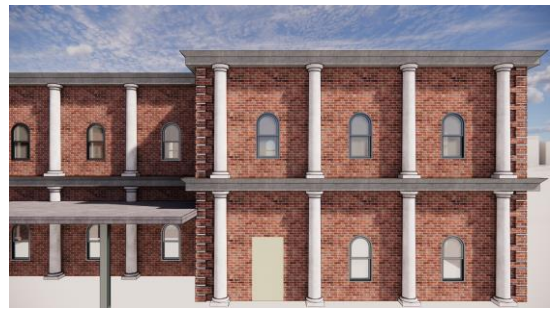

distance between openings = 3 m

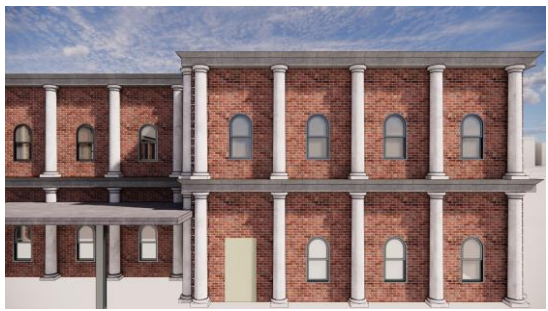

distance between openings = 2.5 m

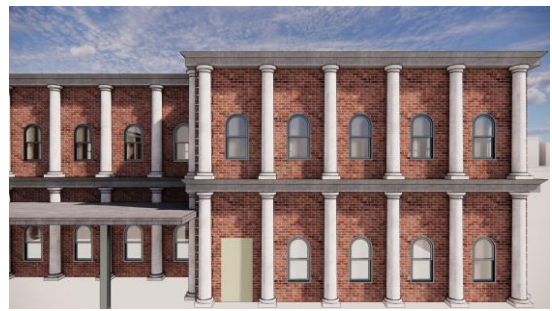

distance between openings = 2 m

**Figure 3.** Different values in meters for the distance between openings parameter are reflected in real-time on the generated BIM model.

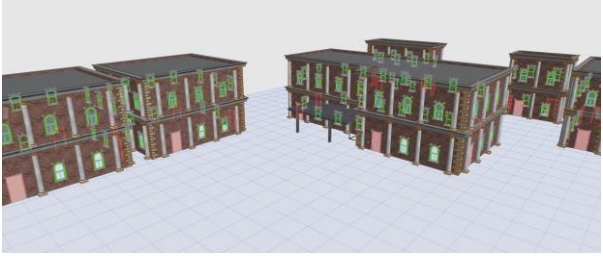

**Figure 4.** Doors placed via their corresponding point lists as selected in the UI (in red).

The approach taken to calculate orientation points is by offsetting wall curves outwards, then projecting the insertion points from the wall curves to the offset curve. This will make all fenestration reveal sides towards the outside. However, using orthogonal projection makes the orientation point lie on the vertical plane that contains the insertion point and become perpendicular to the wall reference line, which will be invalid because the opening direction for the fenestration part cannot be determined. Therefore, the solution is to slightly nudge the points along a vector where this condition does not occur. A fenestration object is selected with a BIM library part parameter that uses the generated BIM wall, insertion point, and orientation point.

If fenestrations follow a known design, customized parameters for placing fenestrations can be added in addition to the standard parameters. Each customized parameter for a window requires an insertion point and corresponding BIM part to be manually defined. Manually placed fenestrations were used to closely simulate the characteristics of a test case for the training dataset.

### 2.1.3. Generating Architectural Details

A wall element is not always a plain surface with a single monochromatic material. Architectural details on walls are abundant in existing buildings. Quoins, cornices, and pilasters are considered as options for embellishing generated 3D BIM models. Selecting different patterns of these detailing options would construct different patterns of architectural style.

Wall polylines curves are offset by a small arbitrary distance to avoid geometry intersections causing visual occlusions. The resulting edges and vertices from the offset operation are used to generate cornices and quoins respectively.

Quoin objects are placed given an insertion point and can be rotated around this point in the XY-axes given an angle. Typically, quoins are on the corners of a building's exterior walls, which will be the vertices of the polylines whose edges define the wall reference lines. A single two-sided quoin object with two faces separated by a 90° angle is placed at each polyline vertex. Instances of a selected quoin object must be correctly placed for any input building footprint shape parameter. The edges of these polylines—representing wall reference lines—can have different directions and angles between connecting edges if they are represented as vectors in a Cartesian coordinate system. Placing quoins on a building whose shape is defined by an arbitrary input polyline without defining the rotation angle of each individual quoin object would produce a result as seen in Figure 5 (a), where (b) is the desired outcome.

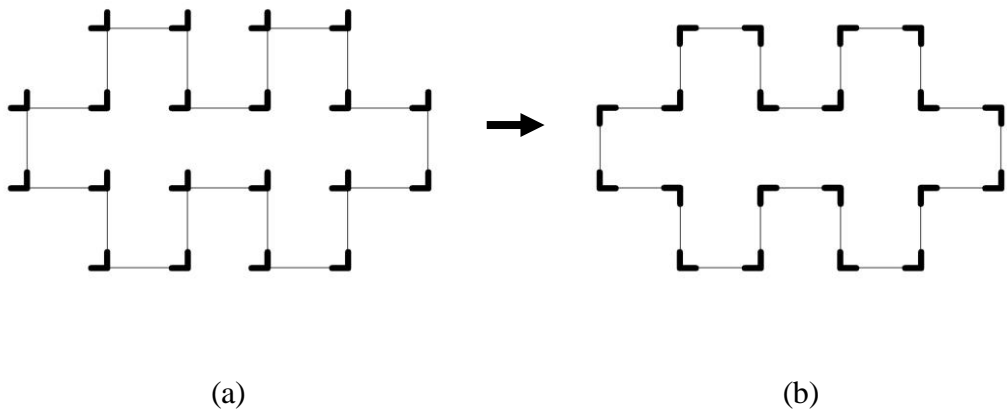

(a)            (b)

**Figure 5.** Quoin rotation angles before (a) and after (b) the correction algorithm is applied.

An algorithm was developed to correctly calculate the quoin rotation angle for each corner. Given $\hat{A}$ and $\hat{B}$ which are the unitized vectors of two adjacent edges, the formula for the algorithm is as follows:

$$x = \theta = \left(arccos\left(\frac{\hat{\imath}\cdot(\hat{A}+\hat{B})}{\|\hat{\imath}\|\,\|\hat{A}+\hat{B}\|}\right)\right)\left(Z \in \hat{A} \times \hat{B} = (X,Y,Z)\right)$$
$$if\ x > 0:$$
$$a = x - 45$$
$$else:$$
$$a = |x| - 225$$

Where: $\hat{\imath}$ is the standard unit vector in the direction of the X-axis, $Z$ is the Cartesian z-component of the cross-product vector $\hat{A} \times \hat{B}$, and $a$ is the calculated angle in degrees for the quoin rotation.

Quoin objects are placed using the wall height parameter, and a quoin style, quoin material, and quoin mortar-color parameters. Cornices are created as walls but use a cornice profile parameter which enables the selection of a profile that can be extruded along a path to create the cornice geometry. Pilasters are generated following the fenestration pattern that was governed by the distance between openings parameter. Increasing the number of windows would increase the number of pilasters and vice versa.

*2.2. Training Data Synthesis*

The training data synthesis method uses BIM models generated from BIMGenE and saved scenes with distinct high dynamic range imaging (HDRI) backgrounds and lighting conditions to create the image pairs of the training dataset. Each generated model inherently contains photorealistic materials and BIM object metadata. Different design typologies can be generated indefinitely; however, user-defined criteria were picked for generating models following the building typology in the test data chosen for the ANN evaluation experiments. The building footprint shape of the test case was the initial criterion for setting the parameters; each generated model uses a variation of the parameters' values following a fixed building footprint shape parameter. Another criterion was introducing custom parameters for windows and doors based on the test case. Figure 6 shows a visual comparison between a photo of the test case and a BIM model automatically generated from parameters that were selected to visually simulate the test case as close as possible. BIMGenE can provide high-fidelity synthetic images that can be either photorealistic yet divergent from real photos or simulate an existing set of photos depending on the breadth of variables chosen.

60 BIM models were generated for creating the dataset. From generation 1 to 30, all buildings are single story. From 31 to 60, buildings are two story. Also, half the generations use customized parameters for windows and doors based on the test case, while the other half use standard parameters to automatically calculate insertion points for fenestrations. Additionally, parameters for different building materials and BIM object parts are defined for each distinct generation. The generated BIM models were exported into FBX file format with embedded materials. To render a specific building generation, a scene is saved where all the other generations are hidden for each generation. The high-quality materials that populate each automatically generated building model enable highly realistic 3D rendering. This replaces the need for manually created 3D models.

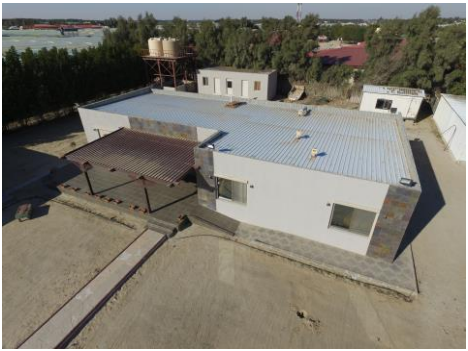 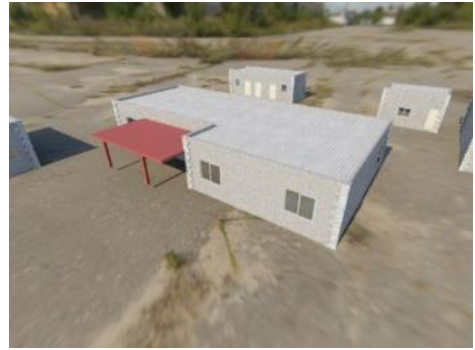

**Figure 6.** Photo of test case vs. realistically-rendered test case BIM model generated from BIMGenE.

*2.2.1 Accounting for Backgrounds and Reflections*

Backgrounds are introduced to teach ANNs to focus on the point of interest (buildings) and block out any noise (i.e., unclear or unwanted objects in the background or foreground). Also, introducing only a single background for the entire training dataset might not make an ANN generalizable when tested using photos with backgrounds that are drastically different from what the ANN was used to. Hence 10 different HDRI backgrounds were used to simulate different realistic conditions. Providing a background class for recognition was deemed important, since ANNs can suffer from failing to provide a variety of recognition objects. This is related to an open problem of ANNs known as the *elephant in the room* problem [30]. To mitigate this problem, ANNs can be shown a large variety of object situations in the training data. Particularly for the application in this research, building backgrounds and surroundings were presumed to have the highest potential risk for the elephant in the room problem. Even if an ANN is highly trained on a variety of building typologies, the same is not true for the countless possible building surroundings and outlier objects that exist in photos. Therefore, the HDRI images were picked to provide variety in the training data.

The HDRI backgrounds were implemented using Open Shading Language (OSL) [31]. Typically, to create a background environment using an HDRI image, the image is projected on a virtual sphere that surrounds a 3D object. However, if there exists a camera that moves around the object (which is the case for the training data), the object will appear disconnected from the environment as if it is floating. To resolve this, a ground projection was applied by flattening the lower half of the virtual sphere; essentially projecting the HDRI image on a hemisphere, where the lower portion of the image is projected on the bottom flat part. The centerpoint of the projected ground was set to the XYZ origin (0,0,0), and the height of the virtual tripod of the HDRI panorama was heuristically set to 15 meters for all the 10 HDRI backgrounds. To simulate shadows on the HDRI backgrounds, a plane object was created with a matte shadow material. This material is non-reflective and non-glossy with a transparent or black alpha map (meaning no texture), but it can receive shadows. This created plane object enables 3D objects to cast shadows on the OSL HDRI backgrounds, which makes the buildings appear more realistically connected to the background environments after rendering.

Glass windows reflect the environment which introduces additional visual complexity for an ANN to understand a window object. Therefore, reflections were simulated by applying the OSL HDRI environments as reflectivity maps for all glass objects in the 60 buildings. This is done for each HRDI environment separately per batch render, so that the glass reflections correspond to the current rendered environment. Depending on the lighting conditions and the location of the camera, the rendered reflections on windows would change in a way that simulates realism similar to photos.

*2.2.2 View Setup*

A moving camera and the method for lighting conditions based on geographic location and time were adapted from previous work [4]. 10 saved scenes of different times of day were used to simulate the virtual sun. The cardinal orientation of the buildings was not modified according to the actual geographic position of the test case, and the orientation/position of generated models from BIMGenE was left as-is, with the virtual sun moving from the negative X direction to the positive X direction in a negative Y incline—simulating the northern hemisphere sun path. The chosen simulated hours of day in the 3D graphics software start from hour 08:00 and continue in subsequent hours until 17:00.

The animation timeline for the camera has 1100 keyframes, where each set of 110 keyframes represents one of the 10 hours of day chosen to simulate the virtual sun. There are 110 distinct camera views. All saved scenes for batch rendering use the same lighting setup provided in the animation timeline. Separate scenes were saved for each single building generation for a total of 60 scenes for batch rendering.

*2.2.3 3D Rendering*

Automatic batch rendering was used to create the dataset. Each batch rendering operation in the 3D software rendered all 60 saved scenes per selected HDRI image for the environment, where each saved scene represents a BIMGenE model generation. The animation timeline that contains keyframes for camera view and time of day was constant for all saved scenes. Meaning, each scene was rendered into 1,100 images, for a total of 66,000 images per batch rendering operation. A physically based rendering software was used. The following is a simple formula that shows the structure of the complete dataset and how the dataset size (number of images) is calculated: #buildings × #views × #hours × #backgrounds = 660,000 images.

Object-ID render passes were produced for each of the 660,000 images rendered using the physically based renderer. The rendered object colors correspond to the BIM object identities in the BIMGenE models. Images were rendered in 341-pixel (width) by 256-pixel (height) sized images. Total rendering time took approximately 2 weeks in a single PC. This duration can be drastically reduced using a professional render farm or cloud computing. Future work can explore real-time rendering methods to reduce the time. Figure 7 shows samples of renderings and corresponding object-ID render pass, including a text description of the BIMGenE generation number, the view number which is the animation timeline frame number (from 1,100 frames), and the number of the HDRI environment.

Both the photorealistic renderings and object-ID render passes were cropped into 256-pixel square images and stitched together automatically to create the complete dataset.

A partial dataset was created by using 12 BIMGenE buildings instead of 60. The buildings are generations 1 to 12. These generations are distinctive by being more similar to the test case than others, since they are all single-story buildings, and have fenestrations that are generated from customized parameters referenced from the test case. By following the previously described formula, the dataset size is 132,000 images. The reason for creating a smaller partial dataset was to compare if an ANN performs better when trained with more epochs using more overfit data versus training with the complete dataset.

## 3. Deep Learning Results and Discussion

Experiments on the test case using training data generated with parametric BIM devised with the methods shown previously were done with two types of ANN models—an image-to-image translation generative adversarial network (GAN) which is Pix2pix by [32], and semantic segmentation convolutional neural networks (CNNs) which are BiSeNet by [33] and PSPNet by [34]. Two training experiments were conducted using Pix2pix with the full dataset (660,000 images) and the partial dataset (132,000 images). Afterward, experiments with the full dataset were conducted using BiSeNet and PSPNet. Table 2 shows the denotations and ANN training parameters for each experiment.

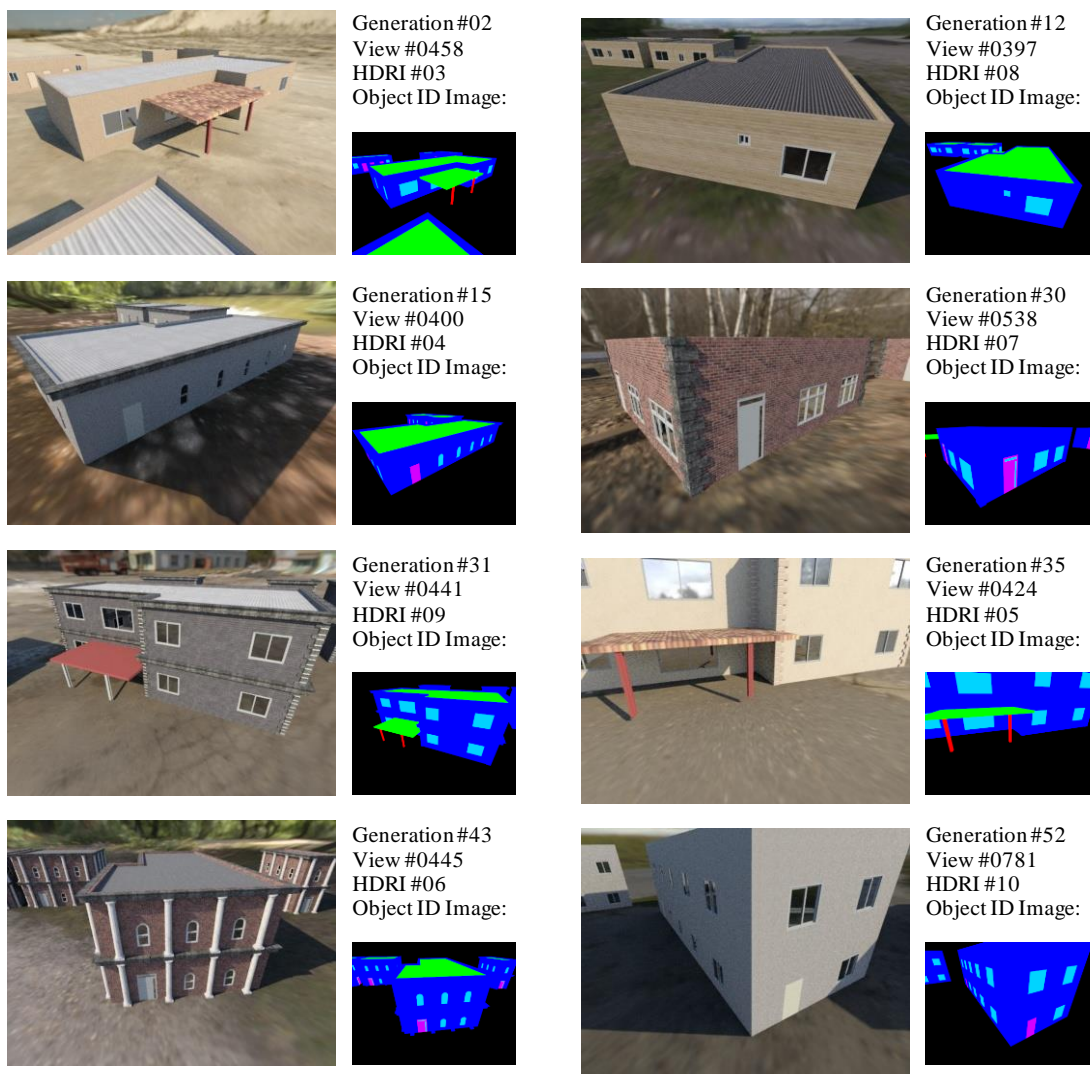

**Figure 7.** Sample renderings and object-ID images for each corresponding BIMGenE generation, view (camera and lighting), and HDRI background.

**Table 2.** Deep learning experiments and training parameters.

| Experiment Denotation | ANN Model | Training | | | |
|---|---|---|---|---|---|
| | | Dataset size | Batch Size | Learning Rate | Augmentation |
| GAN_FULL | Pix2pix | 660,000 | 128 | 0.0002 | Random Flip |
| | | | 32 | | - |
| GAN_PART | Pix2pix | 132,000 | 32 | 0.0002 | - |
| CNN_FULL_1 | BiSeNet | 660,000 | 32 | 0.0001 | - |
| CNN_FULL_2 | PSPNet | 660,000 | 32 | 0.0001 | - |

An evaluation tool was used to evaluate each ANN's semantic segmentation score metrics including mean intersection over union (IoU) score and per-object pixel accuracies (among other standard metrics) on 110 photos of the test case.

**Table 3.** GAN_FULL performance per epoch on test case photoset.

| Input | | Ground Truth | | Epoch 5 | |
|---|---|---|---|---|---|
| Right: Example photo from the testing dataset. | 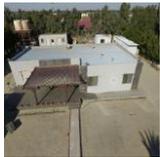 | Right: Example segmentation mask to evaluate from. | 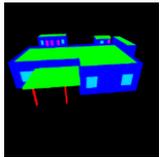 | Avg. Acc. (%) 75.86<br>Window (%) 49.94<br>Door (%) 27.62<br>mIoU 0.385 | 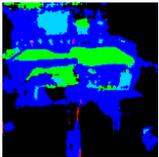 |
| Epoch 10 | | Epoch 15 | | Epoch 20 | |
| Avg. Acc. (%) 73.41<br>Window (%) 58.40<br>Door (%) 25.72<br>mIoU 0.376 | 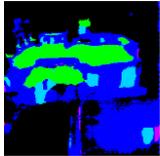 | Avg. Acc. (%) 75.73<br>Window (%) 56.62<br>Door (%) 28.97<br>mIoU 0.400 | 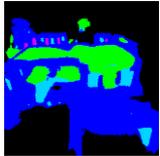 | Avg. Acc. (%) 75.74<br>Window (%) **65.14**<br>Door (%) 26.77<br>mIoU 0.407 | 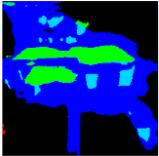 |
| Epoch 25 | | Epoch 30 | | Epoch 35 | |
| Avg. Acc. (%) 75.40<br>Window (%) 63.58<br>Door (%) 26.51<br>mIoU 0.402 | 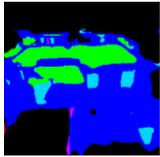 | Avg. Acc. (%) 71.60<br>Window (%) 63.46<br>Door (%) 29.68<br>mIoU 0.377 | 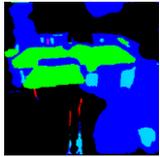 | Avg. Acc. (%) 74.72<br>Window (%) 63.11<br>Door (%) 30.20<br>mIoU 0.403 | 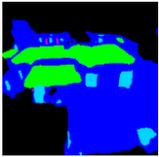 |
| Epoch 40 | | Epoch 45 | | Epoch 50 | |
| Avg. Acc. (%) 72.91<br>Window (%) 60.57<br>Door (%) **30.59**<br>mIoU 0.394 | 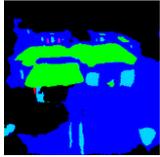 | Avg. Acc. (%) 73.24<br>Window (%) 62.71<br>Door (%) 30.11<br>mIoU 0.402 | 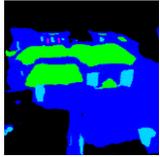 | Avg. Acc. (%) **89.64**<br>Window (%) 59.11<br>Door (%) 29.14<br>mIoU **0.517** | 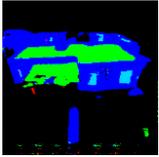 |
| Epoch 55 | | Epoch 60 | | Epoch 65 | |
| Avg. Acc. (%) 83.51<br>Window (%) 60.41<br>Door (%) 26.18<br>mIoU 0.463 | 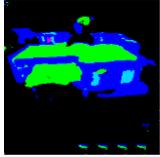 | Avg. Acc. (%) 80.22<br>Window (%) 56.61<br>Door (%) 25.00<br>mIoU 0.406 | 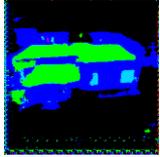 | **GAN FAILURE:** MODE COLLAPSE | 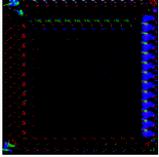 |

*3.1. Training GAN with the Full Dataset*

The first experiment trained Pix2pix on the full dataset for a total of 65 epochs (Table 2). In the first 50 epochs, a batch size of 128 was used. Different batch sizes were tested initially, and 128 was deemed appropriate considering the amount of available time and the capability of the GPU to handle large batch sizes. After epoch 50, training was halted and saved. Training time up to the 50th epoch took approximately 4 days—close to 7,000 seconds per epoch. From epochs 50 to 65, the batch size was reduced to 32, and training was continued. A smaller batch size meant more training time was needed, which was averaging close to 50,000 seconds per epoch.

Table 3 shows evaluation results after every 5 epochs of GAN_FULL on the test case photoset. The training and testing process was erratic; it did not indicate gradual improvement when the saved ANNs were tested on the photos. The results on the 110 test photos were oscillating between improvement and deterioration. For example, the highest mIoU and average accuracy results were in the 50th epoch, the highest window prediction accuracy was in the 20th epoch, and the highest door prediction accuracy was in the 40th epoch (Table 3).

Starting from the 50th epoch, training instability appeared in the form of image artifacts mostly near the borders. Initially, these did not seem to drastically impact the evaluation scores due to the small number of artifacts. Eventually, on the 65th epoch, training became completely unstable assumably due to a known GAN failure called *mode collapse*. Mode collapse is when a GAN gives the same prediction output regardless of the image input [35,36]. The ANN kept producing the same black image with artifacts around the border as shown in Table 3. Mode collapse remains an unsolved open problem affecting GAN training. The ANN trained until the 50th epoch was deemed as the highest-performing one among all epochs following its mIoU score and average accuracy.

**Table 4.** GAN_PART performance per epoch on test case photoset.

| Input | Ground Truth | Epoch 5 | |
|---|---|---|---|
| Right: Example photo from the testing dataset. | Right: Example segmentation mask to evaluate from. | Avg. Acc. (%) **80.87**<br>Window (%) 52.32<br>Door (%) 25.78<br>mIoU **0.401** | |
| Epoch 10 | Epoch 15 | Epoch 20 | |
| Avg. Acc. (%) 80.06<br>Window (%) 53.48<br>Door (%) 26.22<br>mIoU 0.395 | Avg. Acc. (%) 79.51<br>Window (%) 54.75<br>Door (%) 26.47<br>mIoU 0.387 | Avg. Acc. (%) 78.24<br>Window (%) **55.45**<br>Door (%) **26.92**<br>mIoU 0.378 | |
| Epoch 25 | Epoch 30 | Epoch 35 | |
| Avg. Acc. (%) 79.58<br>Window (%) 52.65<br>Door (%) 25.52<br>mIoU 0.388 | Avg. Acc. (%) 77.08<br>Window (%) 48.90<br>Door (%) 25.53<br>mIoU 0.366 | Avg. Acc. (%) 75.33<br>Window (%) 47.48<br>Door (%) 25.12<br>mIoU 0.352 | |
| Epoch 40 | Epoch 45 | Epoch 50 | |
| Avg. Acc. (%) 77.45<br>Window (%) 49.98<br>Door (%) 25.42<br>mIoU 0.371 | Avg. Acc. (%) 76.59<br>Window (%) 50.60<br>Door (%) 25.21<br>mIoU 0.365 | Avg. Acc. (%) 77.19<br>Window (%) 45.69<br>Door (%) 25.02<br>mIoU 0.361 | |
| Epoch 100 | Epoch 150 | Epoch 173 | |
| Avg. Acc. (%) 77.64<br>Window (%) 40.49<br>Door (%) 24.80<br>mIoU 0.360 | Avg. Acc. (%) 78.21<br>Window (%) 43.16<br>Door (%) 24.81<br>mIoU 0.363 | Avg. Acc. (%) 77.98<br>Window (%) 40.22<br>Door (%) 24.96<br>mIoU 0.361 | |

*3.2. Training GAN with the Partial Dataset*

The second experiment trained Pix2pix on the partial dataset for a total of 173 epochs (Table 2). A batch size of 32 was used throughout the training process. This is different from the previous experiment due to the smaller-sized dataset (132,000 images), which enabled using a smaller batch size for more training epochs. Training time was close to 1,500 seconds per epoch.

Similar to the previous experiment, the ANN training and testing process was erratic (Table 4). Also, the results in Table 4 show that training for more epochs beyond the 5 initial epochs did not improve semantic segmentation performance—on the contrary, mIoU and accuracy results became worse in subsequent epochs.

Given that the previous experiment showed a sudden drastic improvement on the 50th epoch (Table 3), in this experiment, training was continued to observe if results improve. On the 20th epoch, window and door prediction accuracy were slightly improved; however, mIoU and average accuracy results decreased (Table 4). Experimentation was halted on the 173rd epoch after no improvement was observed. Based on the evaluation scores in Table 4, the ANN saved on the 5th epoch was deemed as the highest-performing ANN in this experiment.

*3.3. Training with CNNs*

The third and fourth experiments were done using semantic segmentation CNNs as shown in Table 2 (CNN_FULL_1 and CNN_FULL_2 respectively). CNN_FULL_1 trained with the full dataset for only 2 epochs. While CNN_FULL_1 achieved good scores early within these first 2 epochs (compared to Pix2pix), training loss kept increasing gradually with no further improvement. CNN_FULL_1 was biased towards the majority classes (mainly wall and background). Meaning, it is sensitive to class imbalance; it tends to focus learning on classes with more pixel representation while ignoring underrepresented classes.

As for the last training experiment CNN_FULL_2, losses were increasing exponentially during each training session with the full dataset. Also, the ANN was not identifying windows, doors, or roofs. Further training was halted, and the ANN was saved after training for 1 epoch. Evaluation scores were logged for comparison with previous experiments. Window accuracy scored 0% across all 110 test photos since windows appear in every single photo of the test set. For doors and roofs, 100% accuracy was registered only for correctly labeled true negatives, meaning the class does not exist in a particular image.

*3.4. ANN Result Evaluation Measures*

The evaluation method uses the standard metrics for evaluating semantic segmentation. The metrics are the pixel accuracy, precision, recall, F1 score, and the mIoU. The evaluation tool compares the binary object classes per-pixel between the predicted image and the ground truth image. Given any one class (e.g., *wall*): true positive *tp* means the class predicted pixel matches the class ground truth pixel; true negative *tn* means the non-class predicted pixel matches the non-class ground truth pixel; false positive *fp* means the class predicted pixel matches the non-class ground truth pixel; and false negative *fn* means the non-class predicted pixel matches the class ground truth pixel. Pixel accuracy is calculated by the following:

$$accuracy = \frac{tp + tn}{tp + tn + fp + fn}$$

For more in-depth analysis, precision, recall, and F1 are calculated by the following:

$$precision = \frac{tp}{tp + fp}$$
$$recall = \frac{tp}{tp + fn}$$
$$F1 = 2 \cdot \frac{precision \times recall}{precision + recall}$$

The mIoU measures the overlap between ground truth *a* and predicted *b* segmentations and is calculated by the following:

$$IoU(a, b) = \frac{a \cap b}{a \cup b} = \frac{tp}{tp + fp + fn}$$

The main metrics that are often considered are the mIoU and average pixel accuracy. The mIoU is more rigorous than pixel accuracy since it (i.e., the mIoU) measures the overlap (i.e., Jaccard index) as opposed to a pixel-weighted average (i.e., with average pixel accuracy). An mIoU score above 0.5 is generally accepted as a good result. The accuracy, precision, recall, F1, and mIoU results are shown in Tables 5 and 6.

*3.5. Comparison of Results on the Test Case Photos*

The benchmark for evaluating performance on the test case is [4] who trained a GAN ANN with data created from a highly photorealistic photogrammetry 3D mesh and a BIM model of the test case. The previous experiments with parametric BIM data were tested using the same 110 photo set of the test case, which facilitates a comparison between the ANNs.

The GAN method results demonstrated better performance on the testing dataset when compared with the semantic segmentation CNNs. Also, the CNNs had a lower unweighted accuracy score since they performed worse with underrepresented classes (window, door, and column). To compare different GAN results trained on different datasets, Figure 8 shows a visual comparison between all trained GANs.

More interesting is the performance of GAN_FULL trained on the full parametric BIM dataset (dataset with 660,000 images) after 50 epochs. The results visually shown on the corresponding GAN_FULL row in Figure 8 demonstrate comparatively close prediction results with [4] who trained using non-generated training data.

For a more complete comparison, the results of [4] and all ANN training experiments denoted in Table 2 are listed in order from highest to lowest mIoU score in Table 5. The GAN_FULL results, which show the highest-performing ANN trained with parametric BIM-based CGI data, are highlighted in bold text (Table 5).

Training with a larger generated dataset is better than training with a more curated dataset for more epochs—as demonstrated in the comparisons between GAN_FULL and GAN_PART results in Figure 8 and Table 5. It can be concluded that training ANNs with data derived from BIMGenE—represented in the dataset—achieved results that are comparable to the ANN from [4].

These previous results and comparisons demonstrate that deep learning with fully generated CGI data can achieve results that are similar to using manually created CGI. This opens the possibility of using parametric generative programs (like BIMGenE) for creating training datasets for self-learning AI.

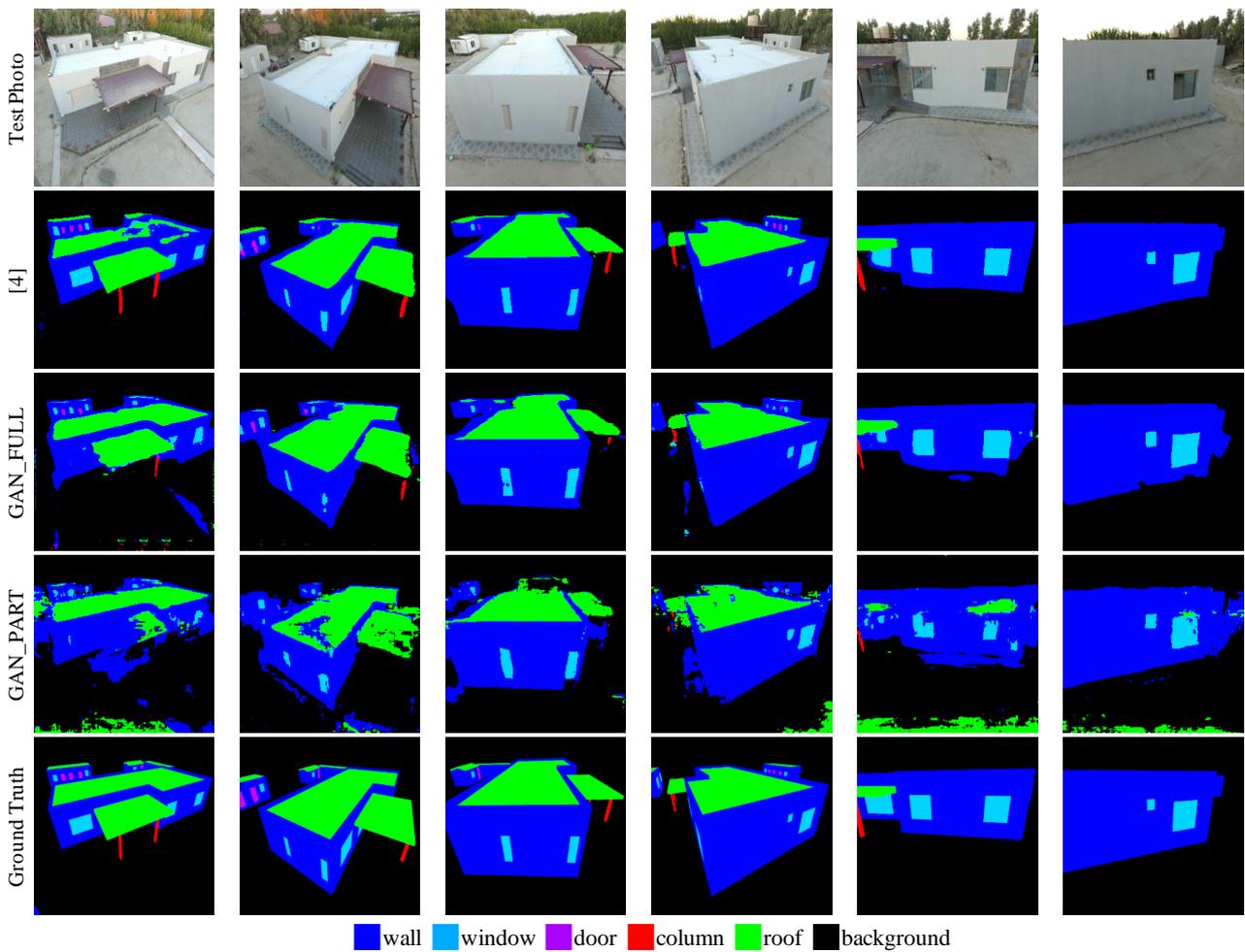

**Figure 8.** Comparison of different ANNs' predicted outputs from test case photo inputs.

**Table 5.** Comparison of evaluation results from different ANNs on test case photoset (listed from highest to lowest mIoU).

| Evaluation Results | Accuracy (%) | Per-Object Accuracies (%) | | | | | | Precision | Recall | F1 Score | mIoU |
|---|---|---|---|---|---|---|---|---|---|---|---|
| | | wall | window | door | column | roof | background | | | | |
| [4] | 92.82 | 93.69 | 85.81 | 55.83 | 88.89 | 76.36 | 94.49 | 0.934 | 0.928 | 0.926 | 0.641 |
| **GAN_FULL** | **89.64** | **86.22** | **59.11** | **29.14** | **69.53** | **85.66** | **92.19** | **0.896** | **0.896** | **0.890** | **0.517** |
| GAN_PART | 80.87 | 81.83 | 52.32 | 25.78 | 59.75 | 70.65 | 83.95 | 0.812 | 0.809 | 0.795 | 0.401 |
| CNN_FULL_1 | 74.89 | 80.51 | 50.39 | 24.84 | 55.45 | 69.51 | 75.60 | 0.764 | 0.749 | 0.730 | 0.373 |
| CNN_FULL_2 | 70.14 | 79.91 | 00.00 | 24.55 | 55.56 | 09.09 | 79.33 | 0.795 | 0.701 | 0.723 | 0.221 |

*3.6. Testing on Arbitrary Building Photos*

The GAN_FULL results demonstrated in the previous comparisons that it is the highest performing ANN that was trained with parametric BIM (from BIMGenE) when tested on the 110 photos of the test case. Since the GAN_FULL ANN has been trained with a large variety of realistic data generated from BIMGenE and rendered with multiple HDRI backgrounds following the method for training data synthesis, it stands to reason to test GAN_FULL on arbitrary building photos besides the 110 test-case photos. Demonstrating that GAN_FULL works well on arbitrary photos brings credence to the usefulness of automatically generated CGI data for deep learning applications on real-world photos.

The testing dataset consisting of arbitrary photos of buildings was downloaded online from Wikimedia Commons, a website that provides public domain and free images under the Creative Commons licensing. A total of 63 photos were downloaded. Then, these images were tested using two trained ANNs; the ANN from [4] and GAN_FULL respectively.

The ANN from [4] performed highly on the 110 test-case photos given the fact that it was trained on photo-like data from a camera-aligned photogrammetry 3D model. In contrast, GAN_FULL performed less; however, since GAN_FULL was trained on a larger domain of data consisting of 660,000 images of 60 generated buildings under different visual situations, it would likely show better performance on the arbitrary photo dataset.

The ground truth color labels need to be manually created for the arbitrary photos to conduct a quantitative semantic segmentation evaluation. Only a fraction of the 63 photos were selected to manually create labels and evaluated using the tool. After testing the arbitrary photo dataset on both ANNs, the 63 results from each ANN (128 total) were visually inspected by eye, and before comparing the results quantitively, 12 of the images were hand-picked from the results of the GAN_FULL ANN based on their quality. Then, another 12 images were randomly sampled from the arbitrary photo dataset. 2 images from the random sample were repeated from the hand-picked set. Afterward, ground truth images were manually created in Photoshop for all images chosen for quantitative testing (12 hand-picked, and 12 randomly sampled).

The quantitative semantic segmentation evaluation results for the arbitrary photos are listed in Table 6. Visual comparisons of the results are shown in Tables 7 and 8 for the hand-picked set, and Tables 9 and 10 for the randomly sampled set. Reasonably, GAN_FULL shows significantly better results on arbitrary photos than the ANN from [4] which scored poorly. In the hand-picked set, the average accuracy and the mIoU for all 12 photos are 86.39% and 0.672 respectively for GAN_FULL, and 58.53% and 0.306 for the ANN from [4]. As for the randomly sampled set, the average accuracy and mIoU are 80.52% and 0.561 respectively for GAN_FULL, and 55.21% and 0.238 for the ANN from [4]. The score differences between the hand-picked set and the randomly sampled set can be attributed to human bias in visually inspecting and selecting the results; however, this disparity is not substantially disparate. A more detailed breakdown of per-object accuracies and other score metrics for each testing set are available in Table 6.

The interpretation of these results demonstrates good generalizability of the parametric BIM-trained ANN on photos of buildings outside of the original test case.

**Table 6.** Evaluation results on arbitrary building photos.

| Evaluation Results | | Accuracy (%) | Per-Object Accuracies (%) | | | | | | Precision | Recall | F1 Score | mIoU |
|---|---|---|---|---|---|---|---|---|---|---|---|---|
| | | | wall | window | door | column | roof | background | | | | |
| Hand-Picked | [4] | 58.53 | 40.79 | 23.73 | 41.67 | 100.0 | 59.70 | 76.57 | 0.633 | 0.585 | 0.585 | 0.306 |
| | GAN_FULL | 86.39 | 85.33 | 88.35 | 68.44 | 100.0 | 71.87 | 91.34 | 0.880 | 0.864 | 0.862 | 0.672 |
| Random | [4] | 55.21 | 42.52 | 19.40 | 58.40 | 75.08 | 23.75 | 73.51 | 0.580 | 0.552 | 0.555 | 0.238 |
| | GAN_FULL | 80.52 | 85.42 | 69.62 | 76.32 | 78.80 | 54.65 | 84.98 | 0.839 | 0.805 | 0.806 | 0.561 |

The GAN_FULL ANN demonstrated good results on various typologies and photo angles of buildings, such as frontal, side, and aerial views. Also, it demonstrated higher window prediction accuracy on the arbitrary photos than on the original test case photos. The window class accuracy score is a relevant indicator to compare since it exists in all images in all testing datasets (e.g., unlike column or door), meaning there are no true negatives for windows that could skew the comparison. It also performed well on a low-light photo (the first-row result in Table 8), likely benefitting from being trained on the low-light renderings from the data synthesis method. Also, the ANN was able to properly predict gabled roofs, even though the buildings generated from BIMGenE lacked such roof typology and only had flat roofs.

One interesting result (second row in Table 8) is the performance of the ANN on an interior photo, where wall, window, and door classes were properly identified, while the floor and ceiling (which do not exist as classes) were identified as background. This is a positive intended consequence of learning from the dataset, where any noise or unclear objects in the environment should be considered as a background object (and labeled in black color). It is interesting to see this learning concept being applied by the AI to an interior photo as well, which is drastically different from the intended application of the ANN on building exteriors.

**Table 7.** Evaluation results on arbitrary building photos (hand-picked). Part 1/2.

| Test Photo | Ground Truth | [4] | GAN_FULL |
|---|---|---|---|
| 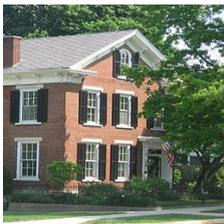 | 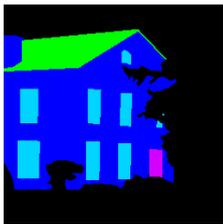 | 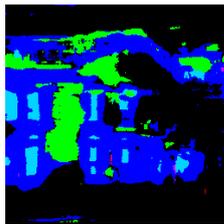 Acc.: 66.65%, mIoU: 0.291 | 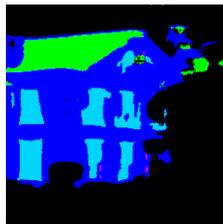 Acc.: 88.12%, mIoU: 0.601 |
| 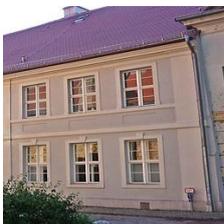 | 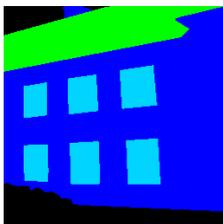 | 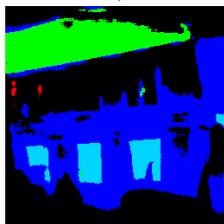 Acc.: 59.33%, mIoU: 0.440 | 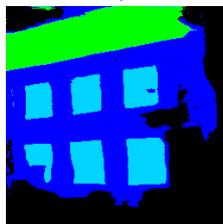 Acc.: 76.38%, mIoU: 0.690 |
| 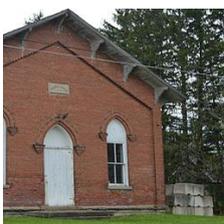 | 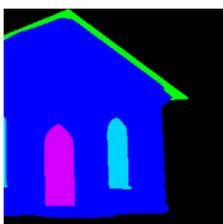 | 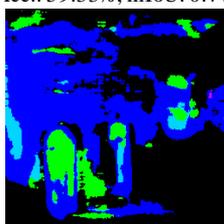 Acc.: 60.80%, mIoU: 0.221 | 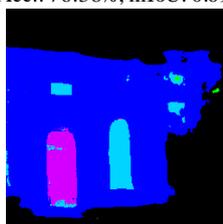 Acc.: 81.38%, mIoU: 0.578 |
| 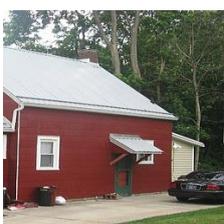 | 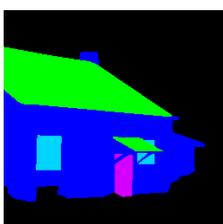 | 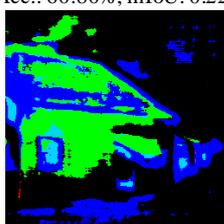 Acc.: 57.30%, mIoU: 0.230 | 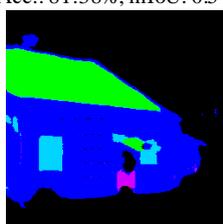 Acc.: 93.06%, mIoU: 0.711 |
| 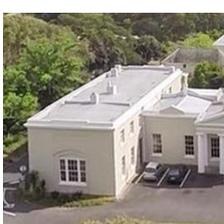 | 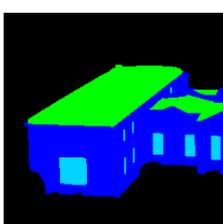 | 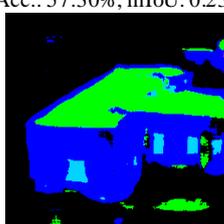 Acc.: 86.98%, mIoU: 0.649 | 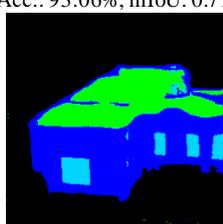 Acc.: 94.37%, mIoU: 0.797 |
| 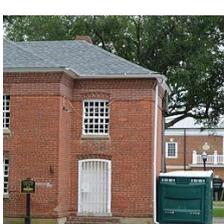 | 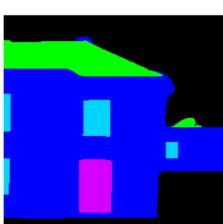 | 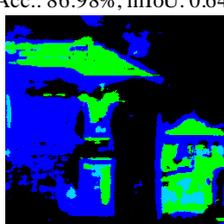 Acc.: 56.48%, mIoU: 0.258 | 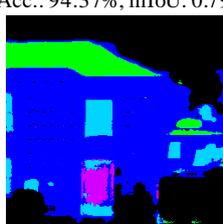 Acc.: 85.09%, mIoU: 0.673 |

■ wall ■ window ■ door ■ column ■ roof ■ background

**Table 8.** Evaluation results on arbitrary building photos (hand-picked). Part 2/2.

| Test Photo | Ground Truth | [4] | GAN_FULL |
|---|---|---|---|
| 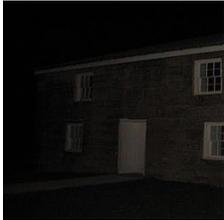 | 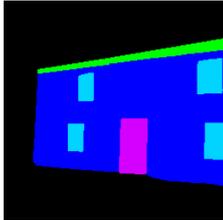 | 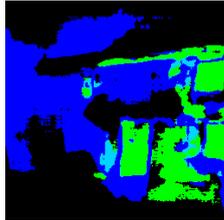 Acc.: 53.58%, mIoU: 0.190 | 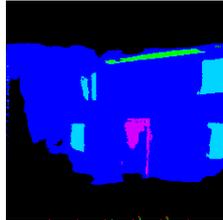 Acc.: 84.89%, mIoU: 0.575 |
| 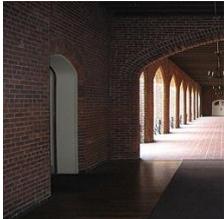 | 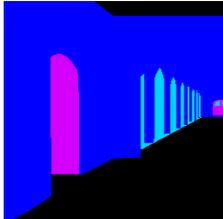 | 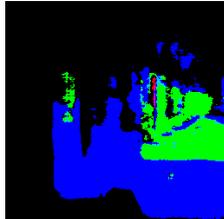 Acc.: 18.31%, mIoU: 0.068 | 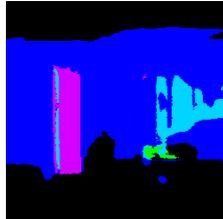 Acc.: 73.62%, mIoU: 0.572 |
| 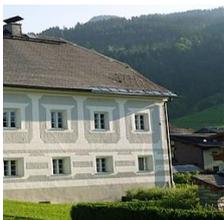 | 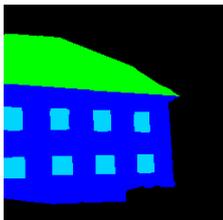 | 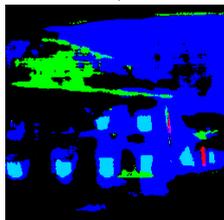 Acc.: 42.80%, mIoU: 0.285 | 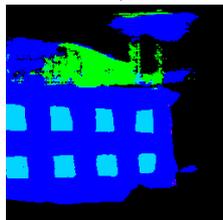 Acc.: 85.87%, mIoU: 0.693 |
| 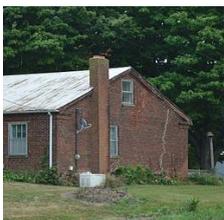 | 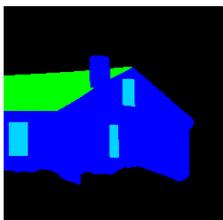 | 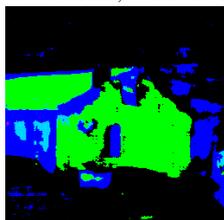 Acc.: 67.28%, mIoU: 0.320 | 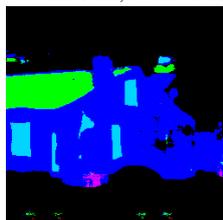 Acc.: 91.18%, mIoU: 0.773 |
| 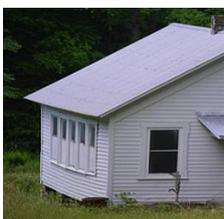 | 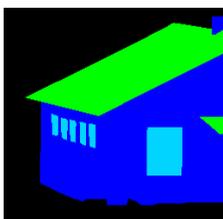 | 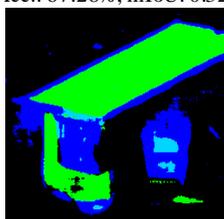 Acc.: 62.70%, mIoU: 0.413 | 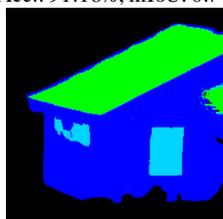 Acc.: 92.32%, mIoU: 0.814 |
| 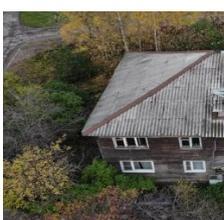 | 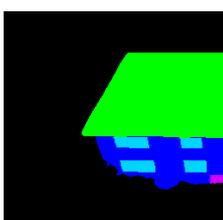 | 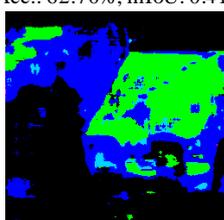 Acc.: 70.16%, mIoU: 0.312 | 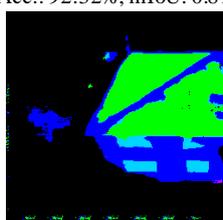 Acc.: 90.42%, mIoU: 0.586 |

■ wall ■ window ■ door ■ column ■ roof ■ background

**Table 9.** Evaluation results on arbitrary building photos (randomly sampled). Part 1/2.

| Test Photo | Ground Truth | [4] | GAN_FULL |
|---|---|---|---|
| 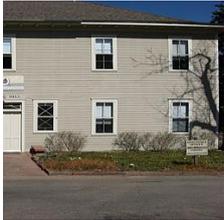 | 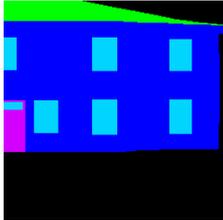 | 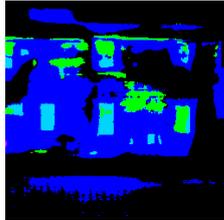 Acc.: 55.87%, mIoU: 0.228 | 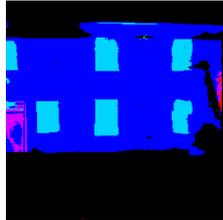 Acc.: 84.95%, mIoU: 0.563 |
| 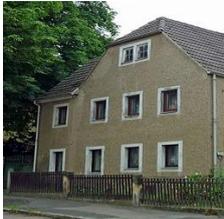 | 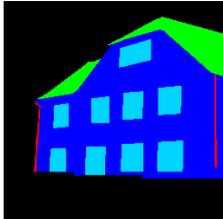 | 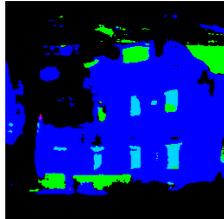 Acc.: 74.39%, mIoU: 0.347 | 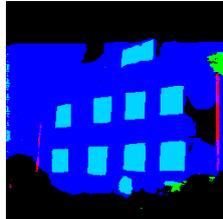 Acc.: 74.13%, mIoU: 0.485 |
| 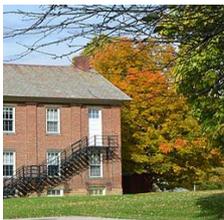 | 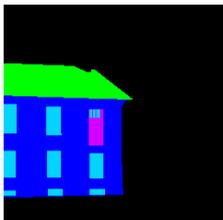 | 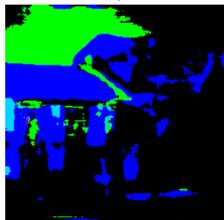 Acc.: 55.51%, mIoU: 0.179 | 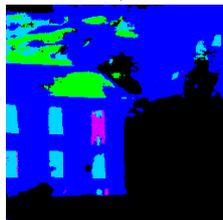 Acc.: 63.21%, mIoU: 0.443 |
| 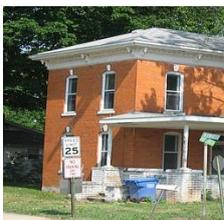 | 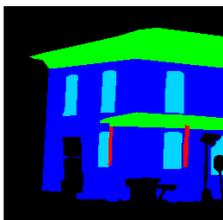 | 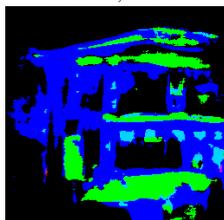 Acc.: 52.29%, mIoU: 0.209 | 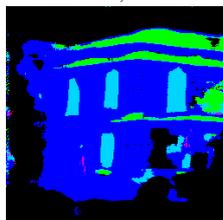 Acc.: 74.28%, mIoU: 0.428 |
| 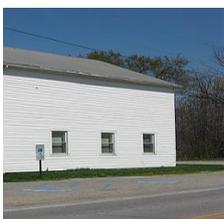 | 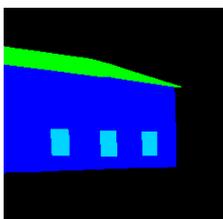 | 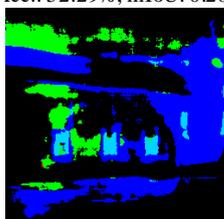 Acc.: 45.61%, mIoU: 0.247 | 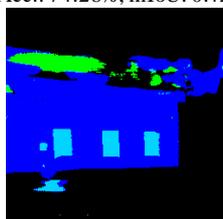 Acc.: 90.76%, mIoU: 0.737 |
| 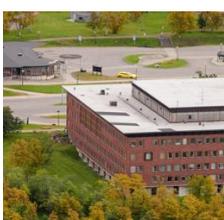 | 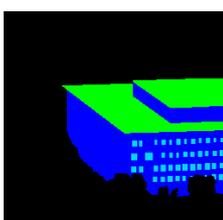 | 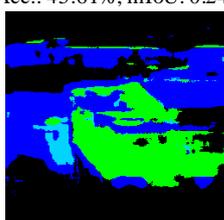 Acc.: 51.66%, mIoU: 0.196 | 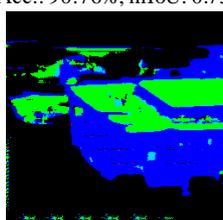 Acc.: 83.64%, mIoU: 0.521 |

■ wall  ■ window  ■ door  ■ column  ■ roof  ■ background

**Table 10.** Evaluation results on arbitrary building photos (randomly sampled). Part 2/2.

| Test Photo | Ground Truth | [4] | GAN_FULL |
|---|---|---|---|
| 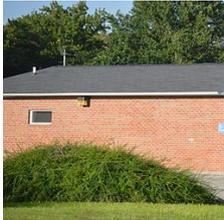 | 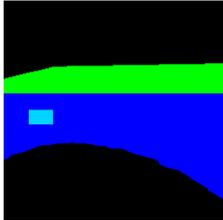 | 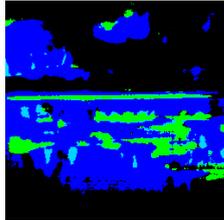 Acc.: 57.26%, mIoU: 0.240 | 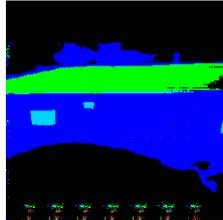 Acc.: 91.63%, mIoU: 0.729 |
| 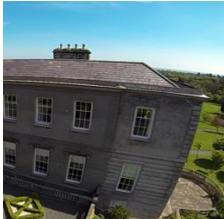 | 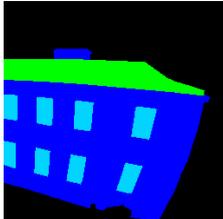 | 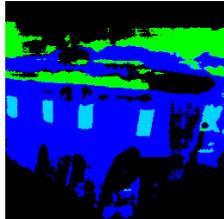 Acc.: 58.42%, mIoU: 0.341 | 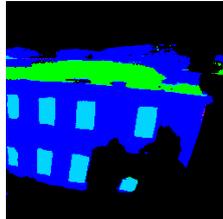 Acc.: 77.71%, mIoU: 0.681 |
| 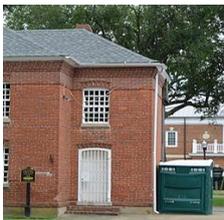 | 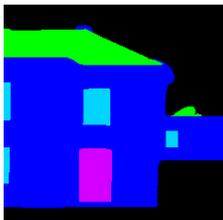 | 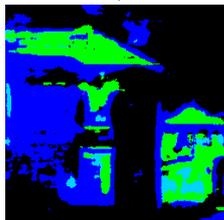 Acc.: 56.48%, mIoU: 0.258 | 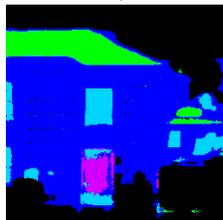 Acc.: 85.09%, mIoU: 0.673 |
| 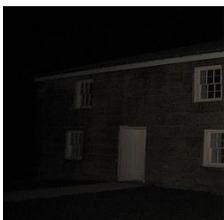 | 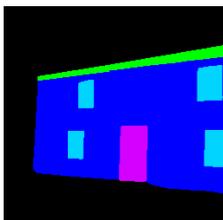 | 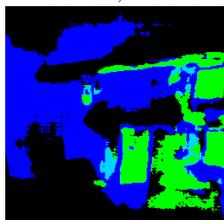 Acc.: 53.58%, mIoU: 0.190 | 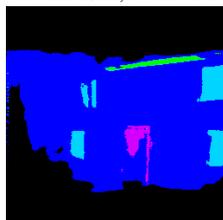 Acc.: 84.89%, mIoU: 0.575 |
| 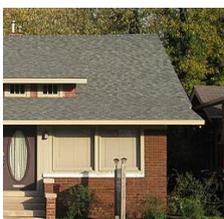 | 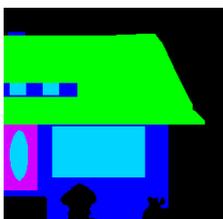 | 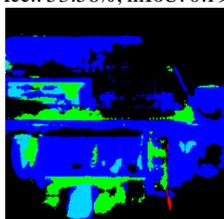 Acc.: 39.76%, mIoU: 0.153 | 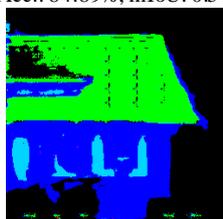 Acc.: 78.87%, mIoU: 0.468 |
| 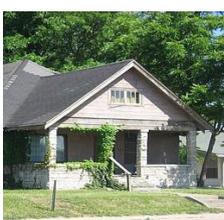 | 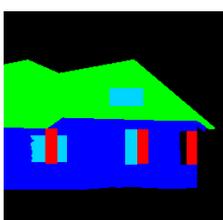 | 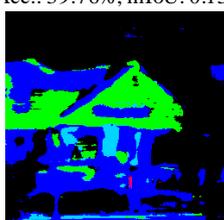 Acc.: 61.63%, mIoU: 0.264 | 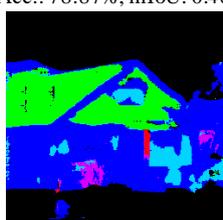 Acc.: 77.08%, mIoU: 0.433 |

■ wall ■ window ■ door ■ column ■ roof ■ background

## 4. Conclusions

Automatically generated BIM models can provide high-quality training data for an ANN to recognize building objects in photos. A machine cannot differentiate between digital images of virtual buildings and real-world photos of buildings. This is advantageous for machine learning to emulate human visual understanding of the built environment. In contrast to machines, we as humans can recognize different variations of objects well after only a few times we see them—that is through our faculties of extrapolation from little information and imagination from memory [37–39]. A machine does not have similar mental faculties; often, if an ANN sees an object that is slightly different than what it is used to, it will fail to recognize it. It will need to be trained by being shown many different examples to be proficient and generalizable on outlier objects. What is suggested from the results is that a machine can be given a faculty akin to imagining buildings via a generative program to assist in its learning by enabling it to explore different variations of building objects and expand its learning domain.

Achieving the desired outcome of building object recognition in photos using training data generated from BIMGenE showed that synthetic training data can be generated instead of needing manually modeled 3D buildings; consequently, deep learning models can have access to virtually unlimited training data. Training ANNs following this approach using parametrically generated BIM models is a novel contribution.

The broad significances of the work are as follows: (1) validating that training data for deep learning can be replaced with synthetic data, (2) proving that BIM and CGI can provide high-quality synthetic training data for ANNs, and (3) showing that training data synthesis can be automated using parametric BIM instead of using pre-existing BIM models and CGI.

Direct applicable benefits to AEC include object recognition for semantic 3D reconstruction of buildings, which is needed for example to automate BEM for green architecture and BIM for construction documentation. The automated construction of semantic 3D building models from 3D scans of existing buildings is desirable especially when considering that constructing them is still a manual process done by experts due to the challenge of automatically extracting building semantics. Automatically generated parametric BIM data can help produce the necessary datasets to train ANNs that are needed for the semantic enrichment of 3D scans.

In the future, parameters can be randomized to generate multiple divergent CGI datasets automatically. A generative approach can be applied to draw the building footprint shape for BIMGenE such as a stochastic recursive method to draw the curves. Also, the presented methods can adapt to improvements in technology: Newer ANNs may improve accuracy, more computational power would enable training with more data, more realistic CGI will train AI better, etc. Future work can develop this research to explore the possibility of a continual learning [40] AI system that can train itself using similar generative programs for recognition tasks. This will be significant for the future of self-learning AI that benefit AEC applications.